\pdfoutput=1

\documentclass[11pt]{article}

\usepackage{EMNLP2022}

\usepackage{times}
\usepackage{latexsym}

\usepackage[T1]{fontenc}

\usepackage[utf8]{inputenc}

\usepackage{microtype}

\usepackage{inconsolata}

\usepackage{times}
\usepackage{latexsym}
\usepackage{xspace}
\usepackage{booktabs} 
\usepackage{graphicx}
\usepackage{subcaption}
\usepackage{amstext}
\usepackage{amsmath}
\usepackage{amsfonts}
\usepackage{amsthm}
\usepackage{enumitem}
\usepackage{bigstrut}
\usepackage{multirow}

\newcommand{\method}{\texttt{PromptEHR}\xspace}

\newcommand{\bx}{\mathbf{x}}
\newcommand{\be}{\mathbf{e}}
\newcommand{\bb}{\mathbf{b}}

\newcommand{\bW}{\mathbf{W}}
\newcommand{\bE}{\mathbf{E}}
\newcommand{\bX}{\mathbf{X}}

\title{\method: Conditional Electronic Healthcare Records Generation with Prompt Learning}

\author{Zifeng Wang and Jimeng Sun \\
University of Illinois Urbana-Champaign \\
\texttt{\{zifengw2,jimeng\}@illinois.edu}
}

\begin{document}
\maketitle
\begin{abstract}
Accessing longitudinal multimodal Electronic Healthcare Records (EHRs) is challenging due to privacy concerns, which hinders the use of ML for healthcare applications. Synthetic EHRs generation bypasses the need to share sensitive real patient records. However, existing methods generate single-modal EHRs by unconditional generation or by longitudinal inference, which falls short of low flexibility and makes unrealistic EHRs. In this work, we propose to formulate EHRs generation as a text-to-text translation task by language models (LMs), which suffices to highly flexible event imputation during generation. We also design prompt learning to control the generation conditioned by numerical and categorical demographic features. We evaluate synthetic EHRs quality by two perplexity measures accounting for their longitudinal pattern (longitudinal imputation perplexity, \texttt{lpl}) and the connections cross modalities  (cross-modality imputation perplexity, \texttt{mpl}). Moreover, we utilize two adversaries: membership and attribute inference attacks for privacy-preserving evaluation. Experiments on MIMIC-III data demonstrate the superiority of our methods on realistic EHRs generation (53.1\% decrease of \texttt{lpl} and 45.3\% decrease of \texttt{mpl} on average compared to the best baselines) with low privacy risks. \footnote{Software is available at \url{https://github.com/RyanWangZf/PromptEHR}.}
\end{abstract}

\section{Introduction}
The prevalence of electronic patient healthcare records fuel the development of machine learning models for many healthcare applications \cite{choi2016doctor,choi2016retain,wang2021online,wang2021lifelong,wang2022survtrace}. However, sharing EHR data usually undergoes strict and expensive de-identification and administration processes thus being difficult. Although there have been attempts on perturbing potentially identifiable attributes as the de-identification step \cite{emam2015anonymising}, they were argued not immune to the hack for re-identification \cite{el2011systematic,choi2017generating}. Alternatively, generating synthetic but realistic EHRs can circumvent data leakage while preserving the patterns of real EHRs for further research and development \cite{biswal2020eva}. 

\begin{figure}[t]
    \centering
    \includegraphics[width=1\linewidth]{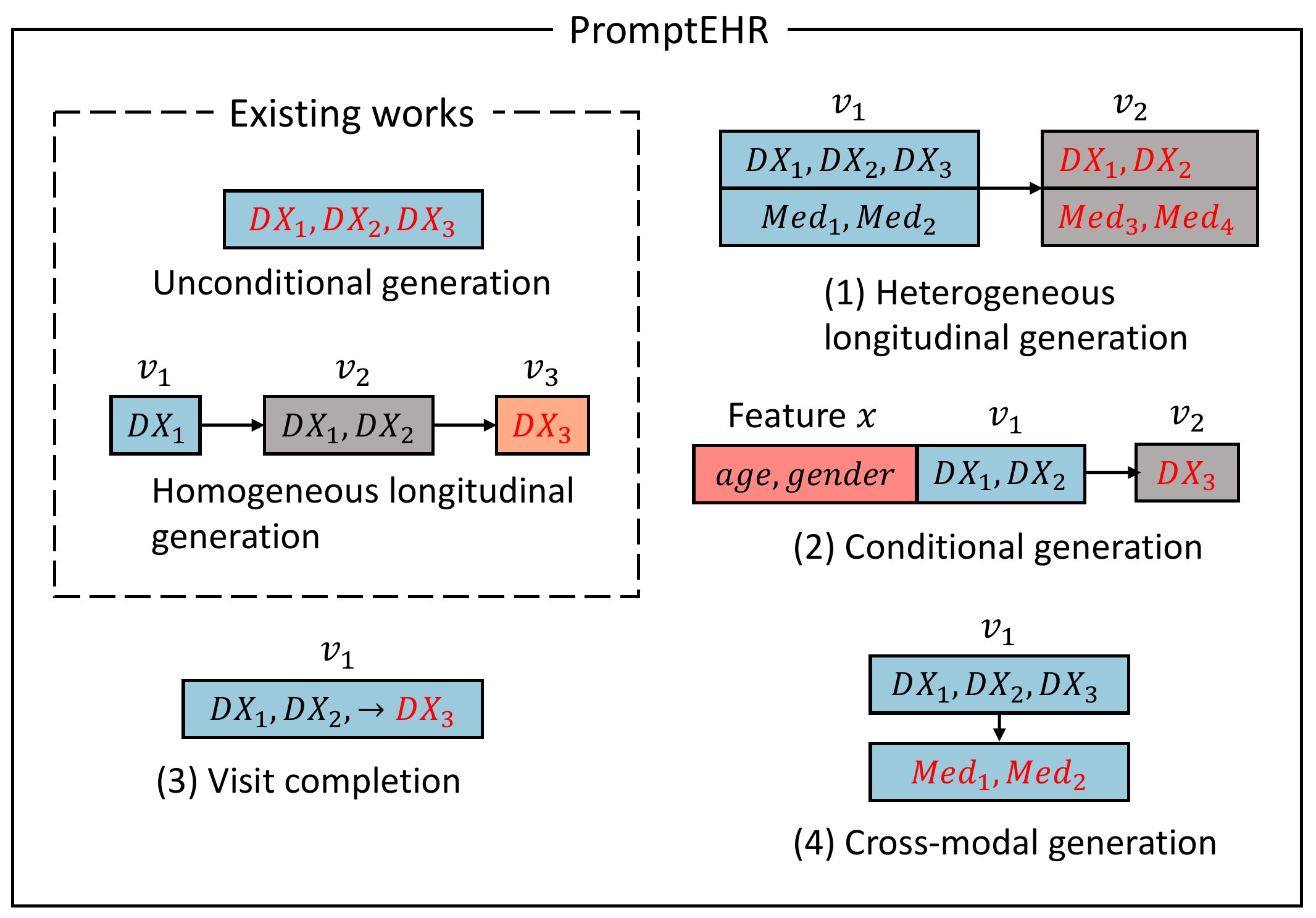}
    \caption{A conceptual demonstration of how \method works more flexible than all existing works. $v_t$ indicates the $t$-th visit; \textit{DX}, \textit{Med} are short for diagnosis and medication events; Red are the targets to generate. Our method (\method) is amenable to four new conditional generation ways thus more controllable and flexible.}
    \label{fig:flexibility}
    \vspace{-1em}
\end{figure}

Deep generative models like GANs \cite{goodfellow2014generative} and VAEs \cite{kingma2013auto} have become popular for unconditional EHRs generation \cite{choi2017generating} and longitudinal EHRs generation \cite{biswal2020eva,zhang2020ensuring} for diagnosis codes. However, EHRs are often multimodal with different types of events, including diagnoses, procedures, medications, and also patient baseline demographic features like age and gender \cite{johnson2016mimic}. GANs \& VAEs usually struggle to model complex multimodal and non-Gaussian distributions as well as sparse one-hot-encoded vectors \cite{xu2019modeling}. By contrast, generative language models (LMs) are proved highly powerful to represent large and complex distributions on discrete data (e.g., texts)  \cite{liu2021swin,radford2021learning}, which makes them promising for EHRs generation.

In this work, we propose to leverage generative language models (LMs) for EHRs generation. We try to generate a sequence of visits with mixed types of events, e.g., diagnosis and medications. As Fig. \ref{fig:flexibility} shows, previous works make unconditional generation for single-modal static EHRs \cite{choi2017generating} or for single-modal longitudinal EHRs \cite{zhang2021synteg}. However, real EHRs are \textit{heterogeneous} with multiple types of temporal events and have baseline patient features, e.g., demographic information. We seek to (1) generate realistic mixed-type longitudinal EHRs with scale and (2) support flexible conditional generation to fit the need for personalized EHRs. Specifically, our contributions are

\begin{itemize}[leftmargin=*, itemsep=0pt, labelsep=5pt]
\item We propose a new EHRs generation method making the best of LMs, which enables generating multimodal EHRs.
\item We design prompt learning for controllable and flexible EHRs generation with LMs.
\item We design comprehensive evaluation for both quality and privacy of the generated EHRs.
\end{itemize}

\begin{figure*}[t]
    \centering
    \includegraphics[width=0.9\linewidth]{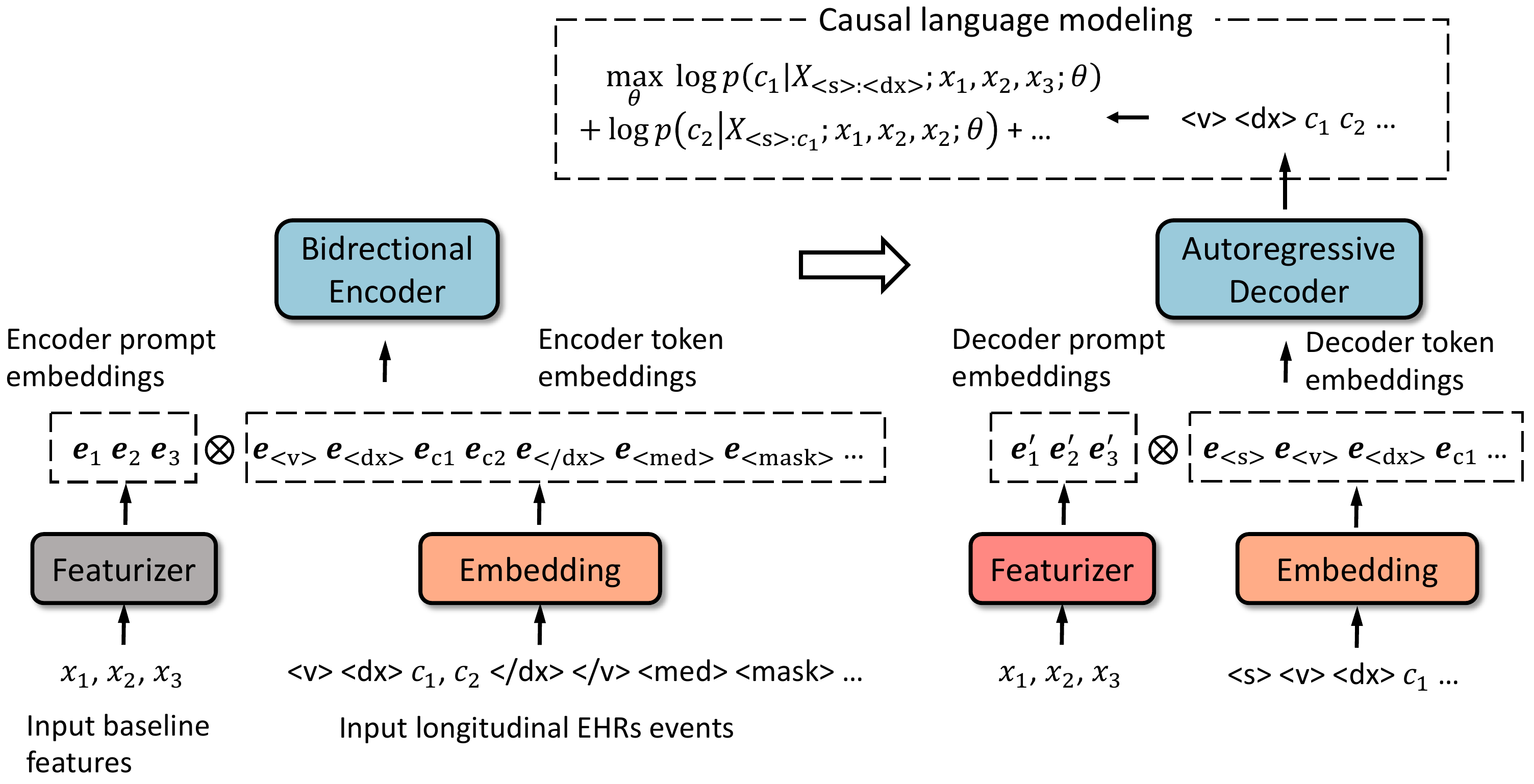}
    \caption{The workflow of \method. The input longitudinal events are transformed to the code sequence by special tokens, e.g., <v> and </v> cover events in the same visit; <dx> and </dx> cover contemporary diagnosis events. Baseline features are encoded to prompt embeddings by two \textit{featurizers} then add to the token embeddings. The model decodes autoregressively and is trained with causal language modeling loss. }
    \label{fig:architecture}
    \vspace{-1em}
\end{figure*}

\section{Related Works}
\subsection{EHRs Generation}
Early works on generating EHRs \cite{lombardo2008ta, buczak2010data, mclachlan2016using} are rule-based methods. However, they were argued not capable of providing realistic data for machine learning tasks and were still vulnerable to re-identification \cite{choi2017generating}. Deep generative models advanced by the power of deep learning, e.g., variational auto-encoders (VAE) \cite{kingma2013auto} and generative adversarial network (GAN) \cite{goodfellow2014generative}, gained most attention recently. \citet{choi2017generating} pioneered in adapting GAN for discrete patient records generation, namely MedGAN, which was followed by improving GANs for EHRs generation \cite{guan2018generation, baowaly2019synthesizing, zhang2020ensuring}; using VAE \cite{biswal2020eva}, hybrid GANs \cite{lee2020generating, cui2020conan}, or conditional GANs \cite{xu2019modeling}. However, most methods only generate static tabular EHRs or longitudinal single-modal EHRs. GANs are often riddled with mode collapse, non-convergence, and instability, which cause their training tricky in practice \cite{saxena2021generative}.
 Moreover, due to the representation limit, GANs struggle in modeling multimodal distributions and sparse one-hot-encoded vectors \cite{xu2019modeling} while EHRs are with these properties. By contrast, we bypass these challenges by LMs. A comprehensive review of EHR synthesis is provided by \citet{wang2022artificial}.

\subsection{Language Models \& Prompt Learning}
LMs are often used for text generation tasks attributed to their \emph{auto-regressive} nature, e.g., T5 \cite{raffel_exploring_2020} and BART \cite{lewis2020bart}. Nonetheless, they cannot be directly applied to EHRs generation since EHRs consist of not only plain clinical notes but also longitudinal sequences of events. Although there were works on encoding and generating medical texts by LMs \cite{amin2020exploring,libbi2021generating,kagawa2021practical,wang2022trial2vec}, none has been done for synthetic EHRs generation. Prompt learning was used to control the topic of text generation \cite{li2021prefix,yu2021attribute,qian2022controllable}. However, they only consider one-hot encoded topics as prefix.  In this work, we leverage prompt learning for EHRs generation conditioned on patient baseline features, which include both categorical and numerical values.

\section{Methods}
In this section, we elaborate on the main framework of \method, including the problem setting, workflow, and training tasks formulation. Next, we discuss the strategies for generating diverse synthetic EHRs with minor loss of quality. Then, we present the recipe proposed for the evaluation for both quality and privacy-preserving ability of the EHRs generation models.

\subsection{Problem Formulation}
Consider there are $N$ patients where the $n$-th patient is represented by 
$X_{n,1:T_n} = \{\bx_n; \bx_{n,1}, \bx_{n,2}, \dots, \bx_{n,T_n}\}$
where $\bx_n$ are the baseline features, e.g., age and gender; $\bx_{n,t}$ signifies events happened at the $t$-th visit; $T_n$ is the total number of visits. For each visit $\bx_{n,t}$, we have $K$ types of events as $\bx_{n,t} = \{\bx_{n,t}^1, \bx_{n,t}^2, \dots, \bx_{n,t}^K\}$. $\bx_{n,t}^k=\{c_1,c_2,\dots,c_l\}$ are all events of type $k$, $l$ is the number of events.

We formulate three basic functions to support EHRs generation:
\begin{itemize}[leftmargin=*, itemsep=0pt, labelsep=5pt]
    \item \textbf{Longitudinal imputation}: given historical visits $X_{n,1:t}=\{\bx_{n,1},\dots,\bx_{n,t}\}$, the model predicts the events in next visit as $\bx_{n,t+1}$;
    \item \textbf{Cross-modality imputation}: given visits with $K-1$ types of events $\bx_{n,t}\setminus \{\bx_{n,t}^k\}$, the model predicts the events belonging to modality $k$;
    \item \textbf{Conditional generation}: given historical visits $X_{n,1:t}$ and the baseline features $\bx_n$, the model makes further predictions.
\end{itemize}
These functions can be combined to synthesize EHRs from the existing partial EHRs with baseline features or from scratch.  

\subsection{Encoding}
The overview is shown by Fig. \ref{fig:architecture}. The first step is to transform the raw inputs $X_{n,1:T_n}$ to token sequences hence acceptable to the encoder.

\textbf{Inputs tokenization.} \method is compatible with all sequence-to-sequence models \cite{cho2014learning}. We choose to utilize BART \cite{lewis2020bart} as the base model. BART uses a bidirectional encoder thus allowing arbitrary corruption for the input sequences and a left-to-right decoder to reconstruct the inputs. Motivated by the application of prompts in language \cite{liu2021pre}, we leverage prompts to specify the inputs. Without loss of generality, we assume two modalities: diagnosis (DX) and medication (Med). Denote \texttt{[X]} and \texttt{[Z]} as the input and answer slots, we can formulate the longitudinal imputation task by a \textit{prefix prompt} problem: \texttt{<v>[X]</v>[Z]}. The model tries to fill the answer slot \texttt{[Z]} which are the events in the next visit; the cross-modal imputation task is built by a \textit{cloze prompt} problem: \texttt{[X]<dx>[Z]} where \texttt{<dx>} signifies the start of diagnosis events and \texttt{[X]} represents the multimodal context events.

\textbf{Conditional prompt featurizer.} We introduce \textit{conditional prompt embeddings} to enable conditional generation based on patient features. We consider both \textit{categorical} $\bx_{cat}$ and \textit{numerical} features $\bx_{num}$. The categorical prompt embeddings $\bE_{cat}$ is obtained by 
\begin{equation}
    \bE_{cat} = (\bx_{cat} \bW_0 + \mathbf{b}) \bW_1.
\end{equation}
$\bx_{cat}$ has $m_c$ mulit-hot encoded indices indicating the classes of each feature; $\bW_0 \in \mathbb{R}^{m_c \times d_0}$; $\bW_1 \in \mathbb{R}^{d_0 \times d_1}$. Therefore, $\be_{cat}$ encodes the instruction of $\bx_{cat}$ and steers the LM to generate specific populations. We transform $\bx_{num} \in \mathbb{R}^{m_u}$ to $\be_{num}$ with another set of $\bW_0$, $\bW_1$, and $\bb$. $\bE_{cat}$ and $\bE_{num}$ then prepend to token embeddings by
\begin{equation} \label{eq:prompt_emb}
    \bE = [\underbrace{\bE_{cat}; \bE_{num};}_{\text{Prompt Embeddings}} \bE_{tok}]
\end{equation}
to serve as the inputs to the encoder. We build the inputs for the decoder with the other featurizer to get $\bE_{cat}^\prime$ and $\bE_{num}^\prime$ and the shared token embeddings $\bE_{tok}$. 

\subsection{Decoding \& Training}
The inputs tokens for the decoder are shifted encoder inputs such that the decoder predicts the next token based on the prior tokens. Denote the context by $\bX$ and the target event by $\bx$, the true conditional distribution is $p(\bx|\bX)$. For instance, in the longitudinal imputation task, the context is the historical record of the patient $\bX_{1:t}$ and the target is the events in the next visit $\bx_{t+1}$. Correspondingly, $p(\bx|\bX;\theta)$ is the prediction made by the model. We use $\tilde{\bX} \sim q(\bX)$ to represent the perturbations added to the context inputs. The training objective is to minimize the negative log-likelihood as
\begin{equation}
    \mathcal{L} = \mathbb{E}_{\bX\sim p(\bX)}\mathbb{E}_{\bx\sim p(\bx|\bX)}\mathbb{E}_{\tilde{\bX}\sim q(\bX)} [-\log p(\bx|\tilde{\bX};\theta)].
\end{equation}
The model is hence pushed to maximize the predicted probability to the true next tokens $\bx$ conditioned by the corrupted inputs $\tilde{\bX}$. 

We apply the following corruptions during training: (1) Token mask, infill, and deletion; (2) Span shuffle and permutation. For (1), we randomly replace multiple tokens with \texttt{<mask>} or delete as length $\sim \text{Poisson}(3)$. For (2), we randomly shuffle the tokens within the same visits and shuffle the modality orders in the same visits.

\subsection{Harmless Randomness in Generation}
Apart from preciseness, the \textit{diversity} of the generated data is also of great importance. \method samples from the conditional distribution by
\begin{equation}
    \bx \sim p(\bx_t | X_{1:t-1};\theta),
\end{equation}
which allows to adjust diversity by many techniques existing in natural language generation literature. For instance, to prevent low probability events, we can apply \textit{top-k} sampling \cite{fan2018hierarchical}. Temperature is also useful to flatten or sharpen the conditional distribution. More advanced methods, e.g., beam search \cite{welleck2019neural} and nucleus sampling \cite{holtzman2019curious} are all available for exploitation by \method, which brings a great potential to achieve higher quality EHRs with diversity. By contrast, GANs \& VAEs depend on sampling random noise vectors to introduce diversity, which is not controllable and usually undermines generation quality.

\subsection{Quality Evaluation}
We provide a recipe to evaluate EHRs generation on two dimensions: \textbf{accuracy} and \textbf{privacy}. For accuracy, we propose to adopt perplexity which is usually used in the text generation task, defined by the exponent of the average negative log-likelihood (NLL) per word \cite{neubig2017neural}: 
\begin{equation} \label{eq:ppl}
 \texttt{ppl} =  e^{-(\log \prod_{l=1}^L p(c_l|c_{1:l-1};\theta))/L},
\end{equation}
where $p(v_l|v_{1:l-1})$ indicates how the model predicts the next word using all previous words as the context; $L$ is the length of the document; $\theta$ is the model parameter. Intuitively, a random predictor will produce \texttt{ppl} that is equal to the cardinality of vocabulary $|\mathcal{C}|$. We hereby adapt it to the longitudinal imputation perplexity (\texttt{lpl}) and cross-modality imputation perplexity (\texttt{mpl}) taking the structure of EHR into account.

\texttt{lpl} takes the \textit{temporal coherence} of the patient visits into account. For instance, chronic diseases like diabetes can cause complications (e.g., heart disease and kidney failure) in the future. Following Eq. \eqref{eq:ppl}, we can write the \texttt{lpl} of a patient's records $X=\{\bx_1,\dots,\bx_T\}$ as
\begin{equation}\label{eq:tpl}
\begin{aligned}
    \texttt{lpl} &= e^{- \sum_{t=1}^T \log  P(\bx_t|\bx_{1:t-1};\theta)/(l_t*T)}
    \\ &=   e^{- \sum_{t=1}^T \sum_{l=1}^{l_t} \log P(v_l|\bx_{1:t-1};\theta)/(l_t *T)}.
\end{aligned}
\end{equation}
Here, $\bx_t = \{c_1,\dots,c_{l_t}\}$ are all events during the $t$-th admission. Inside this admission, concurrent events are independently generated conditioned on previous visits, therefore we can decompose $p(\bx_t|\bx_{1:t-1};\theta)=\prod_{l=1}^{l_t} p(c_l|\bx_{1:t-1};\theta)$ then come to the results.

\texttt{mpl} accounts for the correlations between modalities. For example,  high body temperature in lab test may correspond to fever in diagnosis. We focus on the $t$-th admission where the joint distribution of all $K$ modalities $p(\bx_t^1,\dots,\bx_t^K|\bx_{1:t-1};\theta)$. We can write the NLL here by
\begin{equation}
\begin{aligned}
        \text{NLL}_t & = - \frac1K \sum_{k=1}^K \log p(\bx_t^k|\bx_t^{1:K\setminus k}, \bx_{1:t-1};\theta) \\
        & = -\frac1K \sum_{k=1}^K \frac1{l^k_t} \sum_{l=1}^{l^k_t} \log p(v_l|\bx_t^{1:K\setminus k}, \bx_{1:t-1};\theta),
\end{aligned}
\end{equation}
where $l_t^k$ indicates the number codes belonging the $k$-th modality. Next, we can track all admissions to obtain the final definition of \texttt{mpl} by
\begin{equation}
    \texttt{mpl} = e^{\sum_{t=1}^T \text{NLL}_t / T}.
\end{equation}

\subsection{Privacy Evaluation}
It is crucial to measure the privacy preserving when sharing the synthetic data. We try to evaluate two privacy risks: \textbf{membership inference} and \textbf{attribute inference}. We split the data into the training data $\mathcal{D}_1= \{X_{n,1:T_n}\}_{n=1}^N$ and testing data $\mathcal{D}_2$, and generate synthetic data $\mathcal{D}_S$ with the same length as $\mathcal{D}_1$.

\textbf{Membership Inference.} Attackers would try to infer the membership of the patient records based on the real records they own. We design this adversary based on shadow training \cite{shokri2017membership}. In the first stage, a shadow model $M_{\text{sd}}$ is trained on $\mathcal{D}_S$. It tries to mimic the performance of the generation model in longitudinal inference.

In the second stage, a membership inference dataset is built based on $M_{\text{sd}}(X)$ where $X \in \widetilde{\mathcal{D}}_S \bigcup \mathcal{D}_2$. $\widetilde{\mathcal{D}}_S$ is a subset of $\mathcal{D}_S$ with the same number as $\mathcal{D}_2$. A model $M_{\text{mi}}: \mathbb{Y}_{\texttt{ppl}} \mapsto \{0,1\}$ is trained to differentiate if $X$ comes from $\mathcal{D}_S$ or $\mathcal{D}_2$. We will then evaluate the success rate of $M_{\text{mi}}$ on identifying $X \in \mathcal{D}_1 \bigcup \mathcal{D}_2$. The better the adversary $M_{\text{sd}}(X)$ and $M_{\text{mi}}$ perform on this evaluation, the higher the privacy risk caused by releasing the synthetic EHRs.

\textbf{Attribute Inference.} We build this adversary following \cite{zhang2021synteg}. In this case, attackers hold some incomplete real records where several sensitive attributes are missing.  They would take advantage of the synthetic data to infer these attributes. Besides, attackers also hold the prior knowledge of association between the attributes, i.e., given the incomplete individual records, how probable another code appears in expectation or $P_0 = p(v_l| \{v_1,\dots,v_{l_t}\}_{t=1}^T \setminus v_l)$. With the prior, the attacker will train an attribute imputation model on the synthetic data $\mathcal{D}_S$, i.e., $\hat{P} = p(v_l| \{v_1,\dots,v_{l_t}\}_{t=1}^T \setminus v_l; \theta_{I})$. The attacker then believe the code $v_l$ exists when $\log \hat{P} - \log P_0 \geq \delta$. $\delta$ is a pre-defined threshold. In experiments, we train another attribute imputation model on $\mathcal{D}_1$ to approximate the prior knowledge. We evaluate the success rate of this attack. Besides, we create a control arm where another imputation model is trained on the test set. Comparison between the control and the treatment (imputation model trained on $\mathcal{D}_S$) suffices for an immediate evaluation of the synthetic data's risk level.

\section{Experiments}
In this section, we designed experiments to answer the following questions.
\begin{itemize}[leftmargin=*, itemsep=0pt, labelsep=5pt]
    \item \textbf{Q1.} How well does \method perform for EHRs generation compared with the state-of-the-art methods on generation quality?
    \item \textbf{Q2.} What is the level of privacy risk on membership inference and attribute inference of the generated EHRs by \method?
    \item \textbf{Q3.} Are the synthetic data useful for the secondary use by predictive modeling in practice?
    \item \textbf{Q4.} How is the generation quality of \method influenced by the size of training records?
\end{itemize}

\begin{table}[t]
  \centering
  \caption{Statistics of the used MIMIC-III data.}
\resizebox{0.5\textwidth}{!}{
    \begin{tabular}{ll|ll}
    \toprule
    \textbf{Item}  & \textbf{Number}    & \textbf{Event Type} & \textbf{Number} \bigstrut\\
    \hline
    Patients & 46,520 & Diagnosis  & 1,071 \bigstrut[t]\\
    Total Visits & 58,976 & Drug  & 500 \\
    Total Events & 5,401,961 & Procedure  & 668 \\
    Events per Patient & 116   & Lab Test & 185 \bigstrut[b]\\
    \bottomrule
    \end{tabular}%
    }
  \label{tab:data_stats}%
\end{table}%

\subsection{Experimental Setup}
\noindent \textbf{Dataset.}
\cite{johnson2016mimic} We use MIMIC-III data which has 46k patients' records collected from the intensive care unit. We pick the diagnosis, procedure, drug, and lab test as the target events for generation. All events in the same admission are seen as contemporary. We randomly split the 46,520 patients records into 39,581, 2,301, 4,633 for the train/validation/test set. The data statistics are available in Table \ref{tab:data_stats}.

\noindent \textbf{Baselines.} We compare the following baselines:
\begin{itemize}
    \item LSTM+MLP. This is the baseline that leverages LSTM \cite{hochreiter1997long} to learn the patient state thus extracting the temporal visit patterns. Based on the state embeddings, MLP layers are able to impute the probability of events within the visit or for the next visit.
    \item LSTM+MedGAN \cite{choi2017generating}. The original MedGAN is not able to do conditional generation and temporal inference. Similar to the first baseline, LSTM is used for capturing temporal patterns as the inputs for MedGAN. Then, the generator of MedGAN will try to make conditional generation for records as realistic as possible to fool its discriminator.
    \item SynTEG \cite{zhang2021synteg}. This is one of the most recent EHRs generation methods. It also consists of a state embedding module and a imputation module. It utilizes transformers \cite{vaswani2017attention} for temporal dependency learning and conditional Wasserstein GAN with gradient penalty (WGAN-GP) \cite{arjovsky2017wasserstein,gulrajani2017improved} for event inference.
    \item GPT-2 \cite{radford2019language}. We pick GPT-2 as the LM baseline that only does causal language modeling on EHRs. Then, it is able to do event generation like texts generation.
\end{itemize}

\begin{table*}[t]
  \centering
  \caption{Longitudinal imputation perplexity (\texttt{lpl}) \& cross-modality imputation perplexity (\texttt{mpl}) of models on different kinds of events. Best values are in bold. $\pm$ value indicates the 95\% confidence interval.}
 \resizebox{\textwidth}{!}{
    \begin{tabular}{l|cc|cc|cc|cc}
    \toprule
    Method/Event & \multicolumn{2}{c|}{\textbf{Diagnosis}} & \multicolumn{2}{c|}{\textbf{Procedure}} & \multicolumn{2}{c|}{\textbf{Drug}} & \multicolumn{2}{c}{\textbf{Lab Test}} \\
    perplexity   & \texttt{lpl}   & \texttt{mpl}   & \texttt{lpl}   & \texttt{mpl}   & \texttt{lpl}   & \texttt{mpl}   & \texttt{lpl}   & \texttt{mpl} \\
    \hline
    LSTM+MLP & 125.1 $\pm$ 5.3 & 122.9 $\pm$ 2.0 & 40.3 $\pm$ 1.7  & 43.8 $\pm$ 0.9  & 173.3 $\pm$ 1.9 & 169.5 $\pm$ 0.5 & 68.9 $\pm$ 0.3  & 71.3 $\pm$ 0.5 \\
    LSTM+MedGAN & 169.2 $\pm$ 6.0 & 109.8 $\pm$ 3.1 & 54.4 $\pm$ 2.5  & 40.1 $\pm$ 1.4  & 197.3 $\pm$ 2.5 & 166.7 $\pm$ 0.9 & 76.9 $\pm$ 0.3  & 66.2 $\pm$ 0.2 \\
    SynTEG & 130.4 $\pm$ 4.6 & 130.0 $\pm$ 2.6 & 46.4 $\pm$ 1.8  & 46.2 $\pm$ 1.5  & 175.6 $\pm$ 2.0 & 175.4 $\pm$ 0.9 & 69.5 $\pm$ 0.2  & 69.6 $\pm$ 0.3 \\
    GPT-2 & 121.1 $\pm$ 1.8 & 134.2 $\pm$ 0.9 & 38.7 $\pm$ 0.9  & 48.2 $\pm$ 0.5  & 166.4 $\pm$ 1.8 & 169.6 $\pm$ 0.6 & 69.7 $\pm$ 0.1  & 69.6 $\pm$ 0.1 \\
    \method  &\textbf{65.9 $\pm$ 2.0}  & \textbf{67.7 $\pm$ 0.6}  & \textbf{13.5 $\pm$ 0.8}  & \textbf{10.1 $\pm$ 0.3}  & \textbf{104.7 $\pm$ 1.8} & \textbf{93.7 $\pm$ 0.5}  & \textbf{24.4 $\pm$ 0.1}  & \textbf{50.1 $\pm$ 0.1} \\
    \bottomrule
    \end{tabular}%
    }
  \label{tab:ppl}%
\end{table*}%

\begin{figure}[t]
  \begin{subfigure}[t]{0.23\textwidth}
    \includegraphics[width=\textwidth]{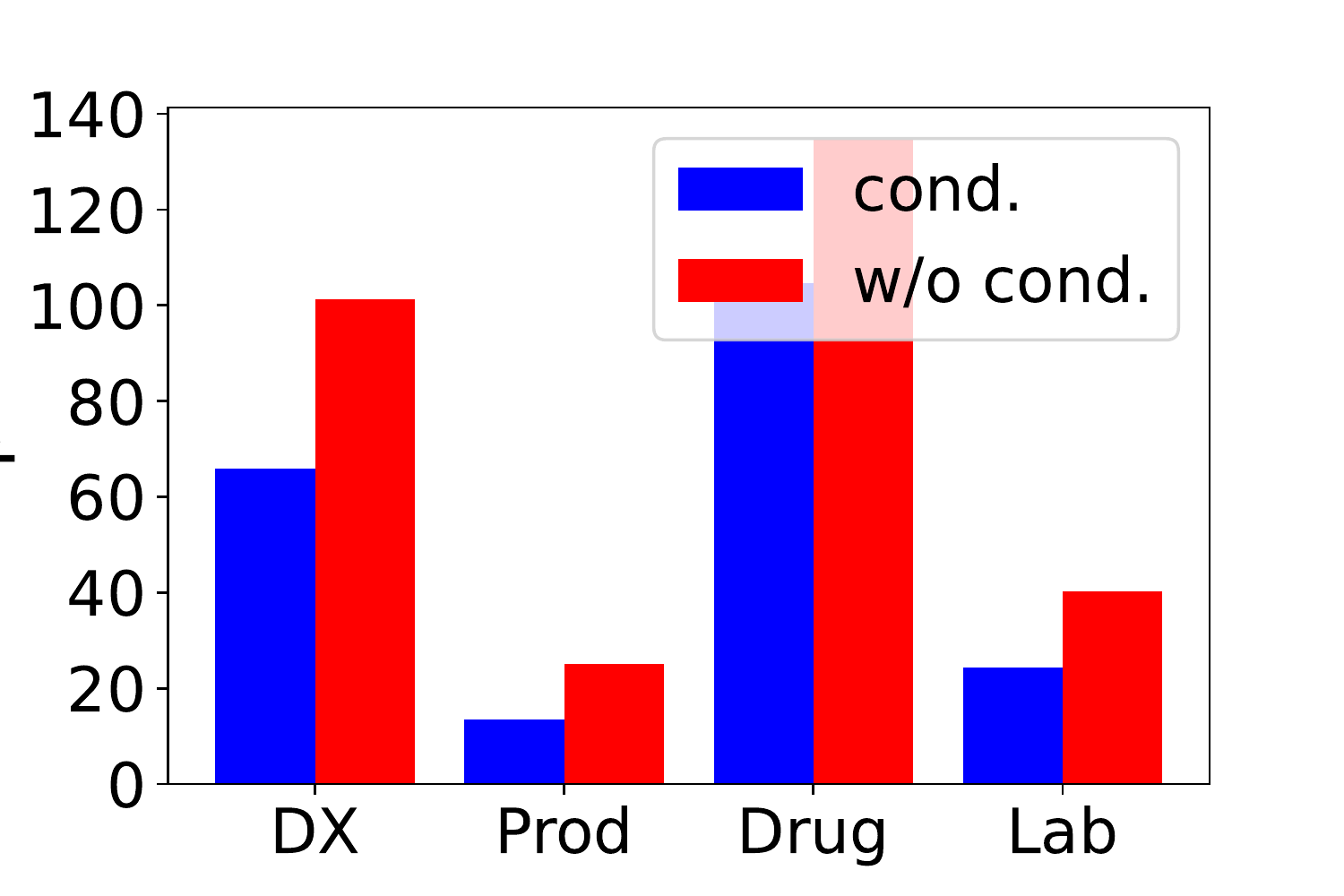}
    \caption{Perplexity (\texttt{lpl}). \label{fig:cond_lpl}}
  \end{subfigure}
  \hfill
  \begin{subfigure}[t]{0.23\textwidth}
    \includegraphics[width=\textwidth]{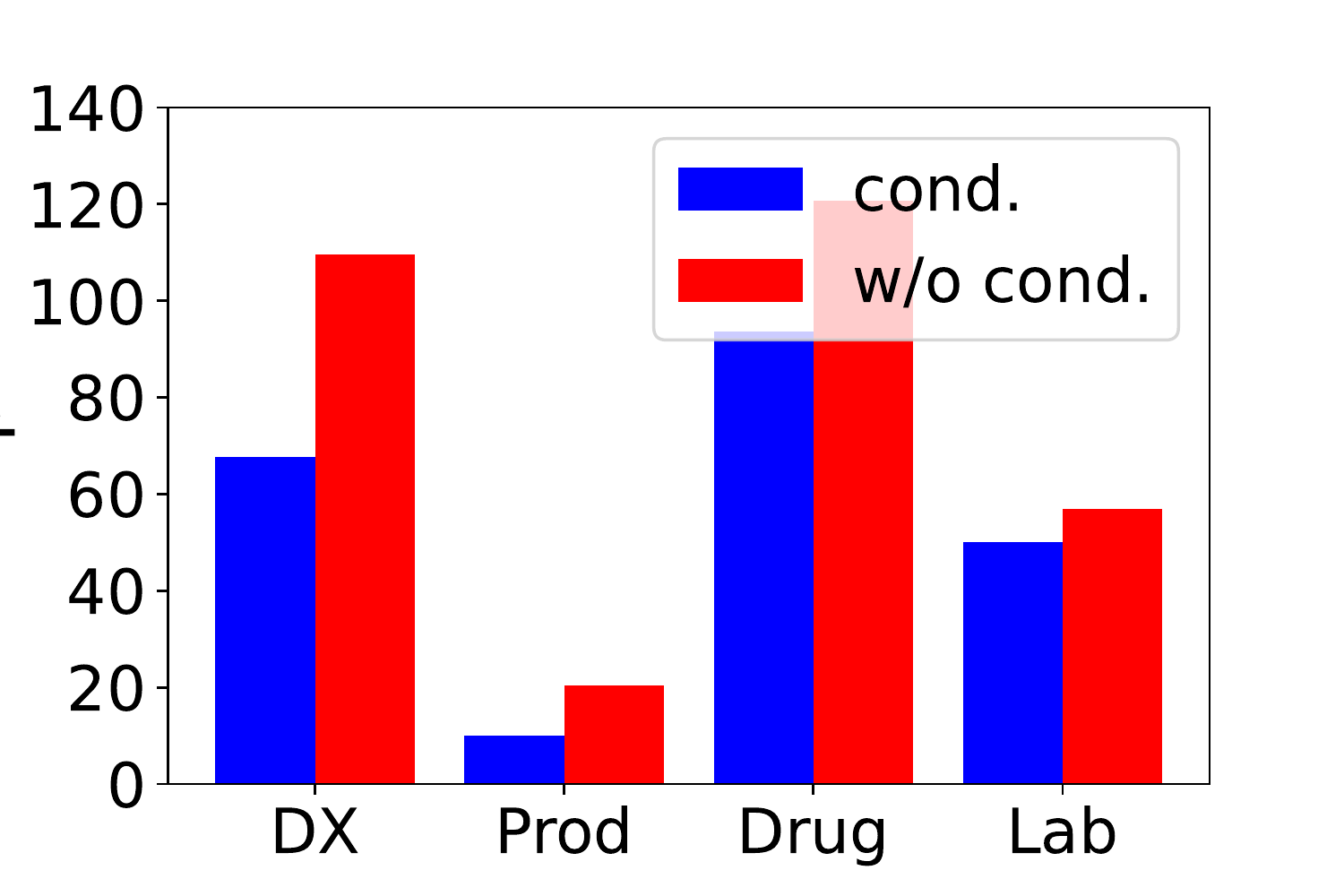}
    \caption{Perplexity (\texttt{mpl}). \label{fig:cond_mpl}}
  \end{subfigure}
\caption{Perplexity compared between generation w/ (cond.) and w/o conditional prompts (w/o cond.) for four types of events. Note that both \texttt{lpl} and \texttt{mpl} are the less the better. \label{fig:cond_lpl_mpl}}
\end{figure}

\begin{figure}[t]
  \begin{subfigure}[t]{0.23\textwidth}
    \includegraphics[width=\textwidth]{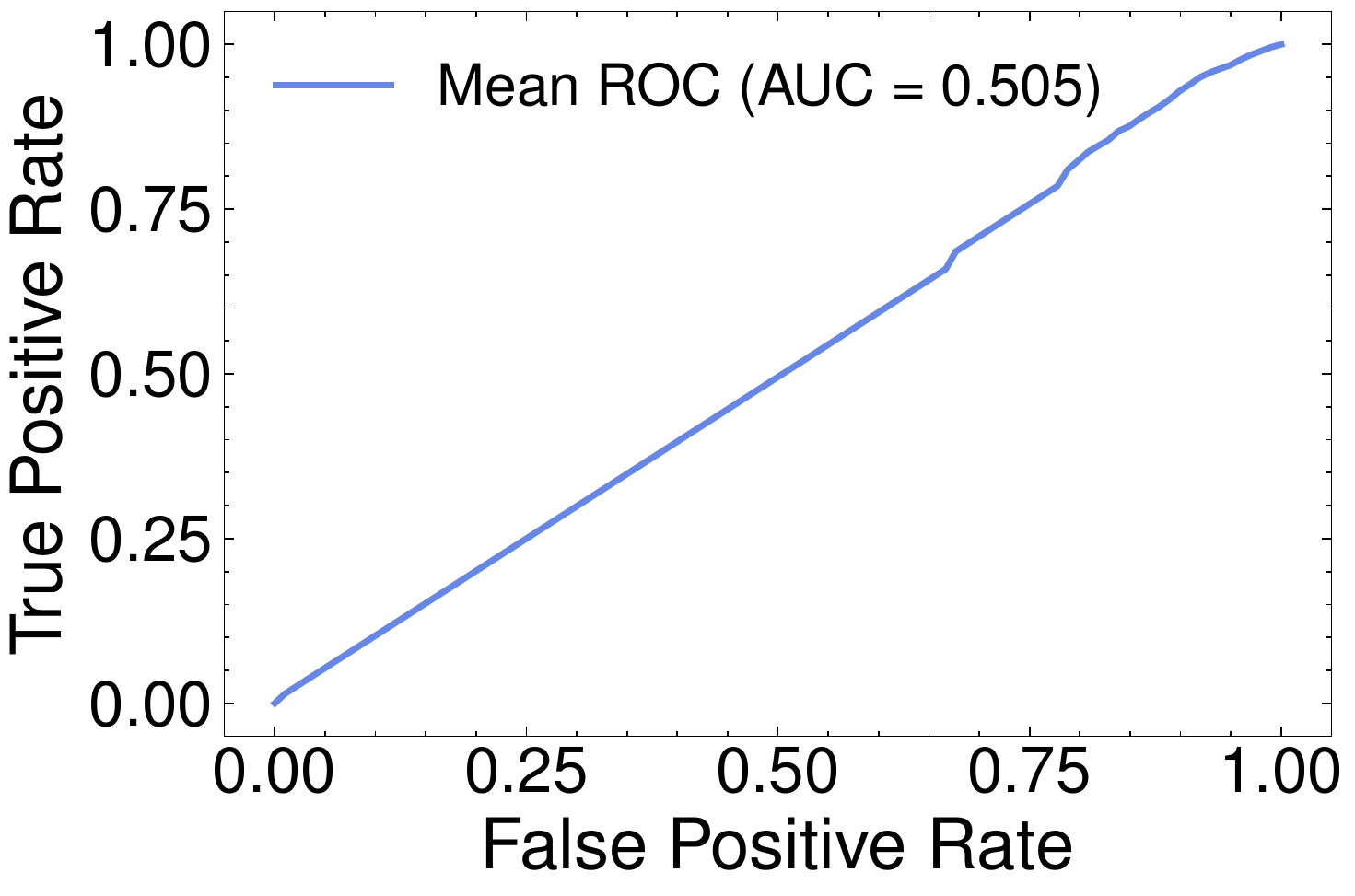}
    \caption{The ROC curve of the membership inference attack by shadow training. \label{fig:mi_adv}}
  \end{subfigure}
  \hfill
  \begin{subfigure}[t]{0.23\textwidth}
    \includegraphics[width=\textwidth]{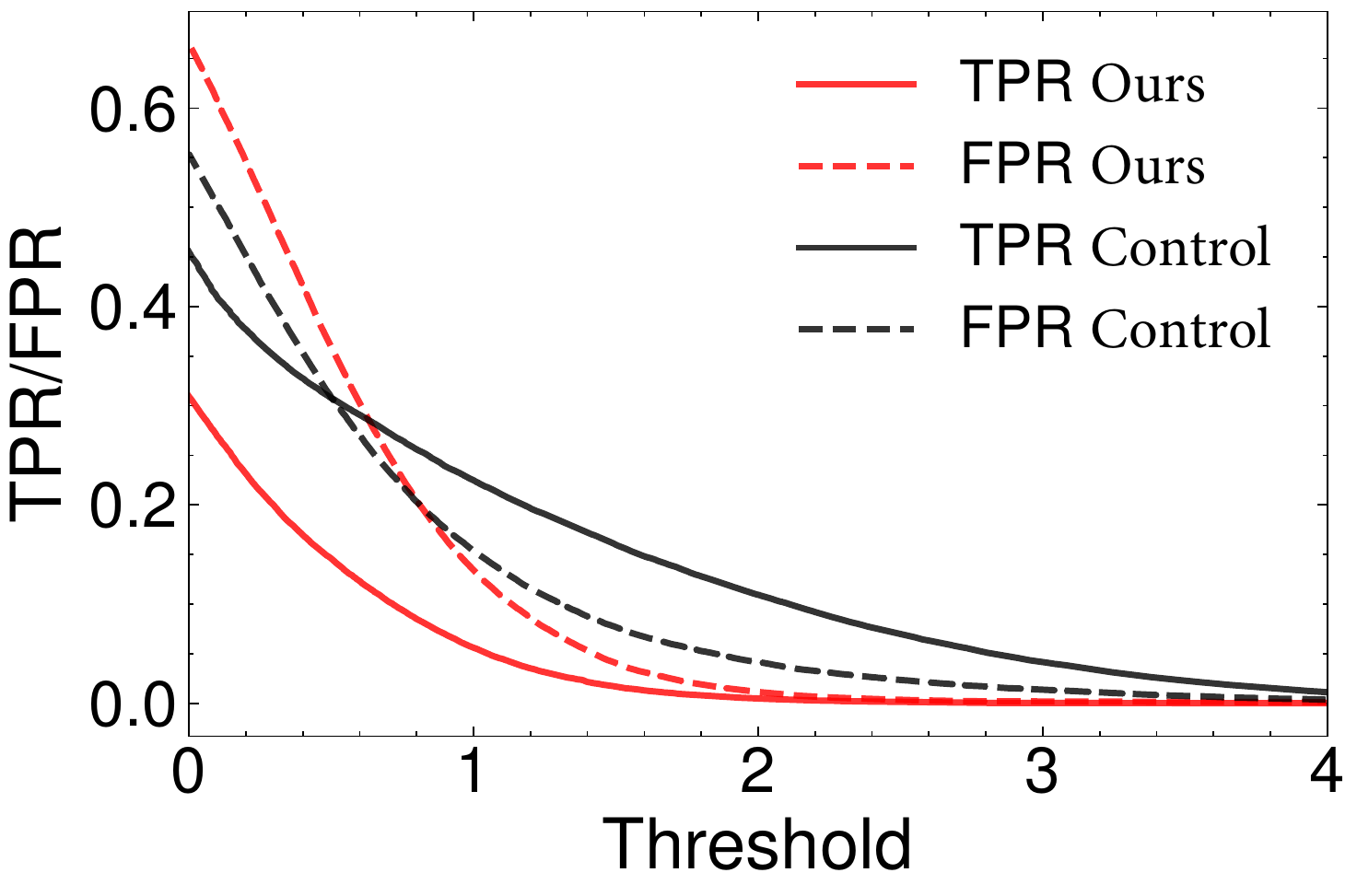}
    \caption{The true positive rate (TPR) and false positive rate (FPR) of the attribute inference attack w.r.t. different thresholds $\delta$. \label{fig:ai_adv}}
  \end{subfigure}
\caption{Privacy-preserving evaluation on membership inference (left) and attribute inference (right) adversaries. On the right, the \method curves indicate the results of attribute inference model trained on the synthetic data $\mathcal{D}_S$ by \method; the Control curves indicate the one trained on test set $\mathcal{D}_2$. \label{fig:privacy_eval}}
\vspace{-1em}
\end{figure}

\subsubsection{Evaluation metrics}
We use the proposed \texttt{lpl} and \texttt{mpl} to evaluate generation quality. Since perplexity of different patient records vary significantly, we take the median of perplexity across patients for the sake of stability of the performance estimate.

We use two adversaries: membership inference (MI) and attribute inference (AI), to test the privacy risk. In MI, we use LSTM+MLP as the shadow model to mimic the outputs of \method. A three-layer MLP predicts the membership. ROC curve is plotted to evaluate the attack success rate; In AI, we train an LSTM+MLP on $\mathcal{D}_1$ to approximate the prior and another LSTM+MLP on $\mathcal{D}_S$ as the attribute imputation model.

To test the utility of the synthetic data for downstream predictive tasks, we train LSTM+MLP on $\mathcal{D}_S$ or $\mathcal{D}_2$ and test it on $\mathcal{D}_2$ to compute the recall@20/30.

\begin{figure}[t]
  \begin{subfigure}[t]{0.23\textwidth}
    \includegraphics[width=\textwidth]{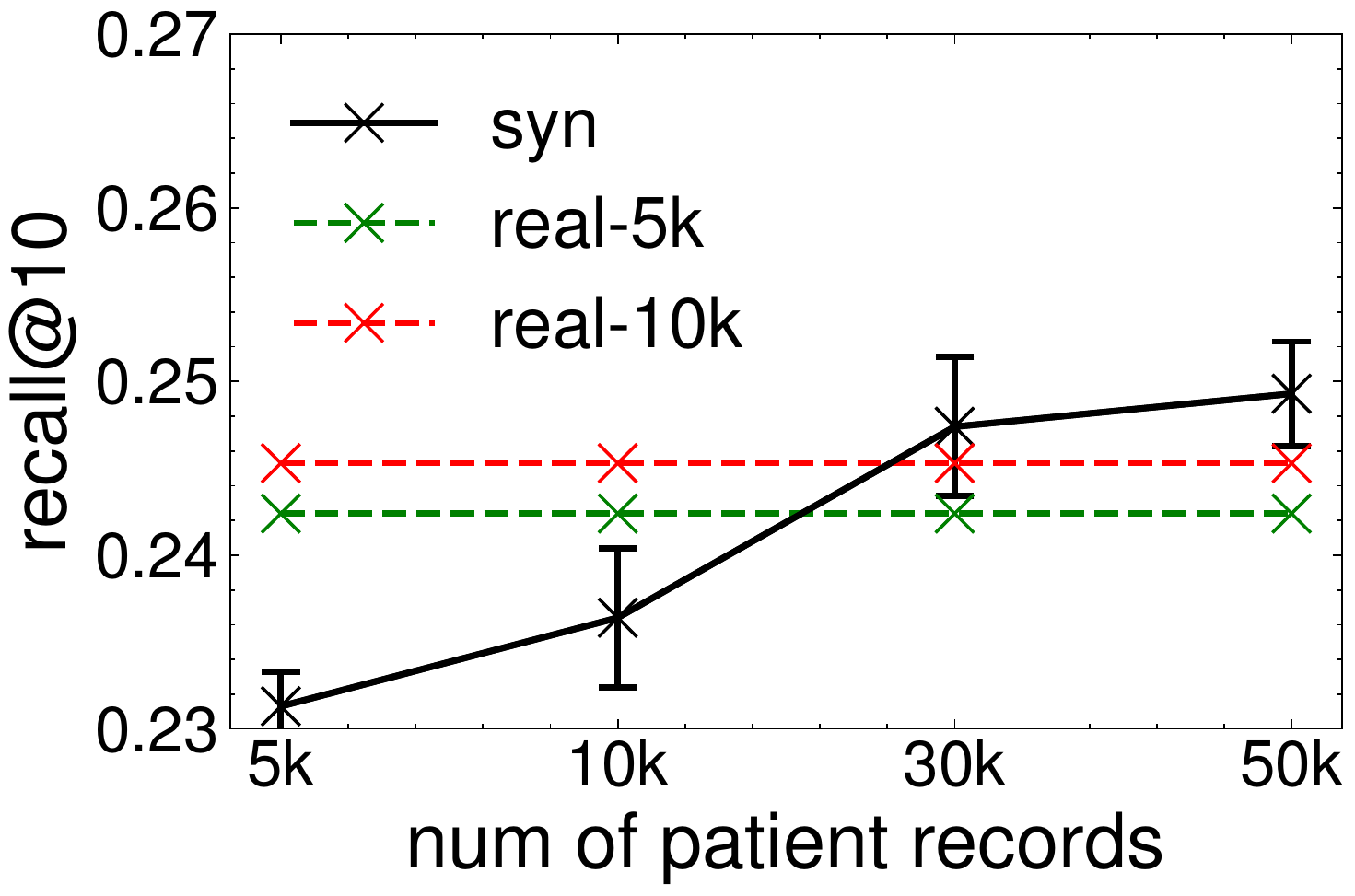}
    \caption{Recall@10. \label{fig:syn_rec_10}}
  \end{subfigure}
  \hfill
  \begin{subfigure}[t]{0.23\textwidth}
    \includegraphics[width=\textwidth]{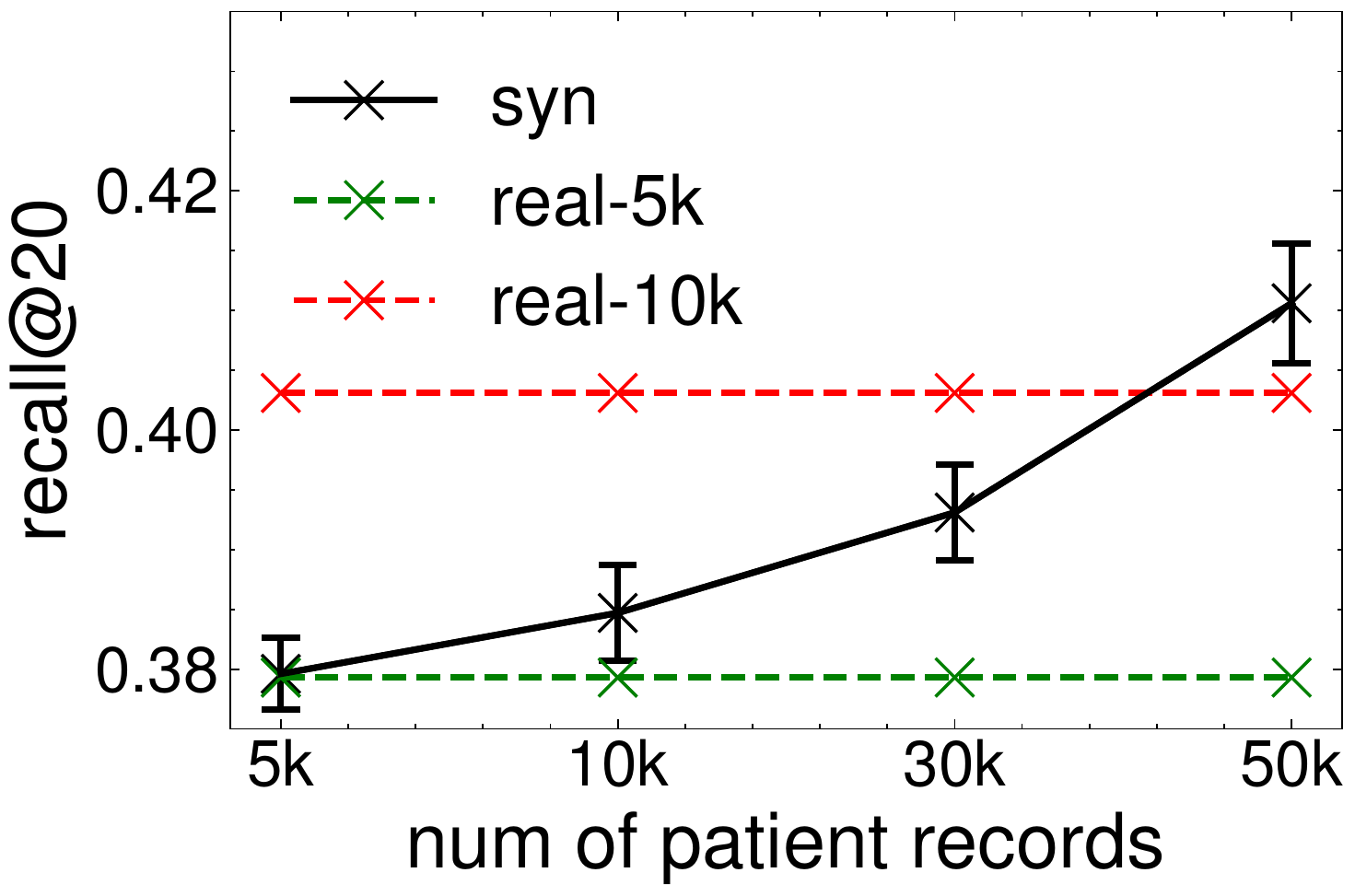}
    \caption{Recall@20. \label{fig:syn_rec_20}}
  \end{subfigure}
\caption{Recall@10/20 of the predictive model on the test set with varying input data size: \emph{syn} indicates the model trained on \emph{fully synthetic data}; \emph{real-5k/10k} indicate trained on 5k/10k real data. Error bars show the 95\% confidence interval which also appear in the following figures. \label{fig:syn_rec}}
\vspace{-1em}
\end{figure}

\subsection{Implementation Details}
All the used LSTM+MLP model consists of a three-layer bi-directional LSTM with 128 hidden dimensions with one 256-dim MLP layer. It is trained with 1e-4 learning rate by Adam optimizer \cite{kingma2014adam}. The 12-layer transformer based pre-trained GPT-2 is trained with 1e-5 learning rate and 1e-4 weight decay by Adam.  We follow the architecture and training protocol from the original papers of MedGAN and SynTEG. 

For \method, we use BART model as the backbone \cite{lewis2020bart}. We use Adam by setting learning rate as 1e-5, weight decay as 1e-4, batch size as 16. The total training epoch is 50 where the first 3 epochs are warm-up steps. During the training stage, the perplexity computed on the validation set is used to pick the best checkpoint. All experiments are conducted with an RTX-3090 GPU, 251 GB RAM, and AMD Ryzen Threadripper 3970X 32-core CPU.

\subsection{Q1. Generation Quality}
The calculated \texttt{mpl} and \texttt{lpl} of all show in Table \ref{tab:ppl}. It is witnessed that \method obtains the best result among all methods. On the contrary, LSTM+MedGAN and SynTEG do not gain better test perplexity than the basic LSTM+MLP. The main reason is that their GAN part takes a noise input except for the learned temporal state embeddings to make conditional generation. GPT-2 works better than LSTM+MLP on temporal perplexity crediting to its power in capturing series pattern through transformers.

Most methods obtain better \texttt{mpl} than \texttt{lpl}. It is intuitive because models know the additional in-visit information from the other modalities for the target modality imputation, thus making better predictions. However, GPT-2 performs worse on \texttt{mpl} than on \texttt{lpl}.
GPT-2 is trained by causal language modeling task where it models the sequence autoregressively. Without the prompt design, it is confused by the order of events within the same visit, which induces deteriorating performance.

Fig. \ref{fig:cond_lpl_mpl} demonstrates the comparison made between generation w/ and w/o conditional prompts for \method. We identify that conditional prompts significantly improve the generation quality as they provide important characteristics of the patients. We are hence able to generate for specific populations with input prompts.

\begin{figure}[t]
  \begin{subfigure}[t]{0.23\textwidth}
    \includegraphics[width=\textwidth]{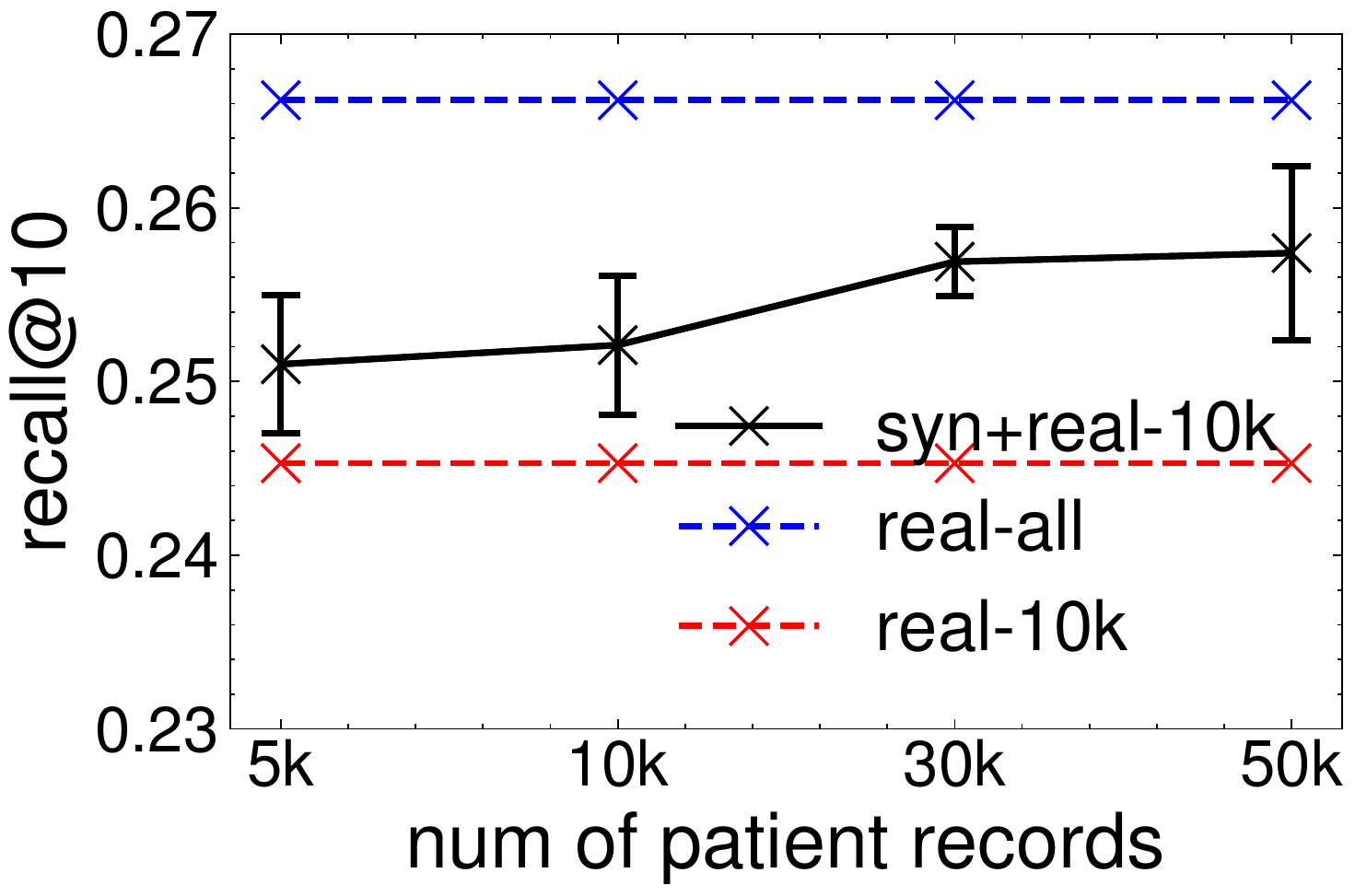}
    \caption{Recall@10. \label{fig:syn_real_10k_rec_10}}
  \end{subfigure}
  \hfill
  \begin{subfigure}[t]{0.23\textwidth}
    \includegraphics[width=\textwidth]{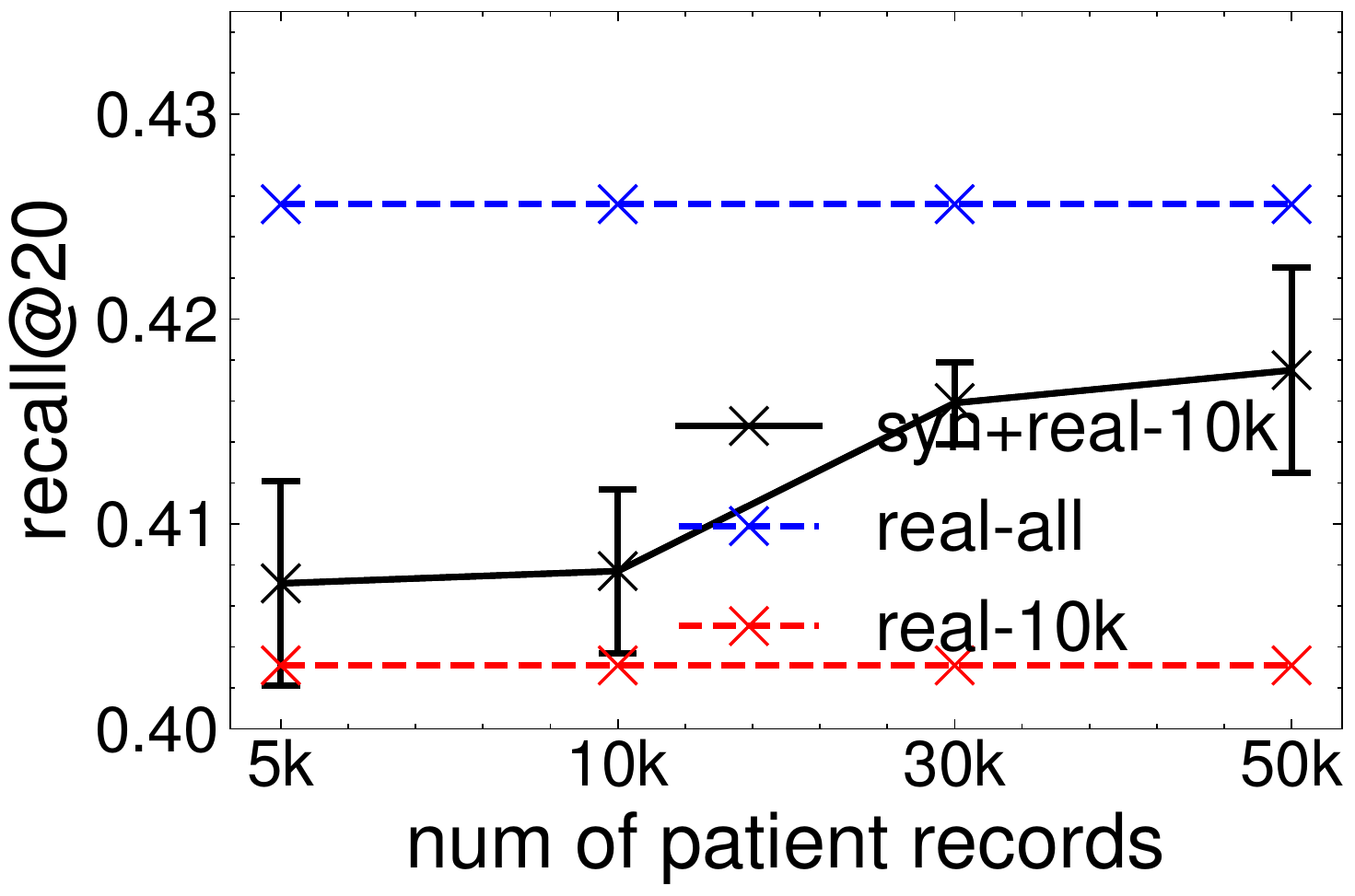}
    \caption{Recall@20. \label{fig:syn_real_10k_rec_20}}
  \end{subfigure}
\caption{Recall of the predictive model on the test set with varying input data size: \emph{syn+real-10k} indicates the model trained on the \emph{hybrid of synthetic \& 10k real data}; \emph{real-10k/all} indicate trained on 10k/all real data. \label{fig:syn_real_10k}}
\vspace{-1em}
\end{figure}

\subsection{Q2. Privacy Evaluation}
We test the privacy preserving ability of the generated synthetic EHRs by applying membership and attribute inference attacks. Results are illustrated by Fig. \ref{fig:privacy_eval}. Fig. \ref{fig:mi_adv} demonstrates the ROC curve consisting of true positive rate (TPR) and false positive rate (FPR) of the membership inference on $\mathcal{D}_1 \bigcup \mathcal{D}_2$. It clearly shows the MI model has bad performance that is near random guess (AUC $\simeq$ 0.5), which means the MI attack gains no sensitive membership information when trained on the synthetic data $\mathcal{D}_S$. 

Fig. \ref{fig:ai_adv} shows the TPR/FPR of attribute inference attack based on shadow training with the varying threshold $\delta$. Here, we cut the curve where $\delta=4$ because all the remaining curves are approaching zero on its right. The threshold $\delta$ adjusts to the confidence level of the attacker, i.e., the smaller $\delta$ is set, the higher probability that the AI is correct we believe. When $\delta=0$, so long as the AI inference probability $P(v_l)$ is larger than the prior $P_0(v_l)$, the AI model will believe the attribute $v_l$ exists. In this scenario, both two models have a high FPR of around 0.6, but the TPR of \method is only near half of the control model. The TPR then keeps a much lower level when $\delta$ increases, which implies the low attribute leakage risk of the synthetic data generated by \method. Although the FPR becomes smaller than Control when $\delta>0.8$, the TPR of \method is approaching zero after that. That means, being conservative for \method avoids inferring some wrong attributes but loses the ability to specify the right attributes at the same time. In a nutshell, the synthetic data by \method has a low risk to leak the attribute information.

\begin{figure}[t]
  \begin{subfigure}[t]{0.23\textwidth}
    \includegraphics[width=\textwidth]{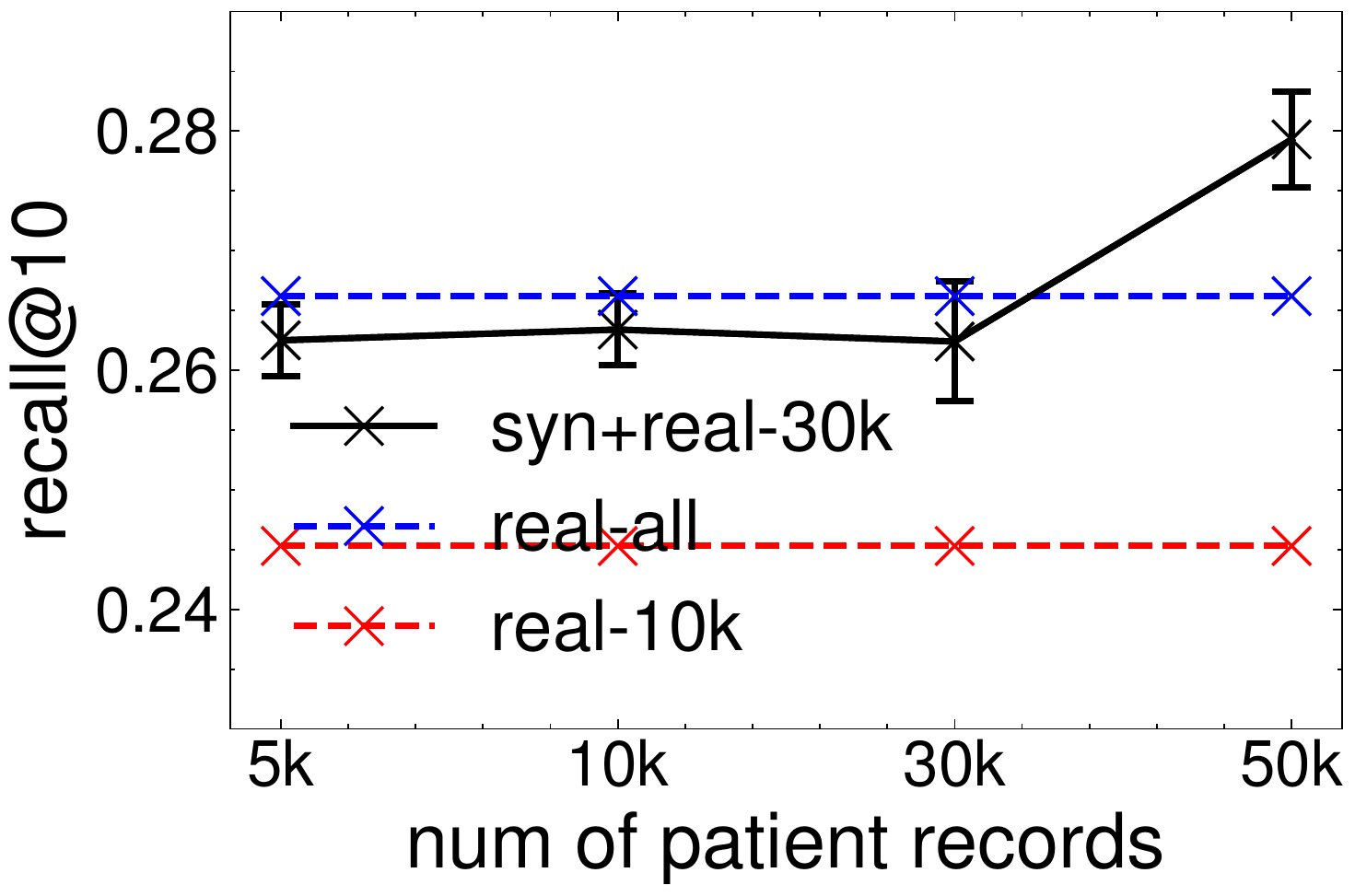}
    \caption{Recall@10. \label{fig:syn_real_30k_rec_10}}
  \end{subfigure}
  \hfill
  \begin{subfigure}[t]{0.23\textwidth}
    \includegraphics[width=\textwidth]{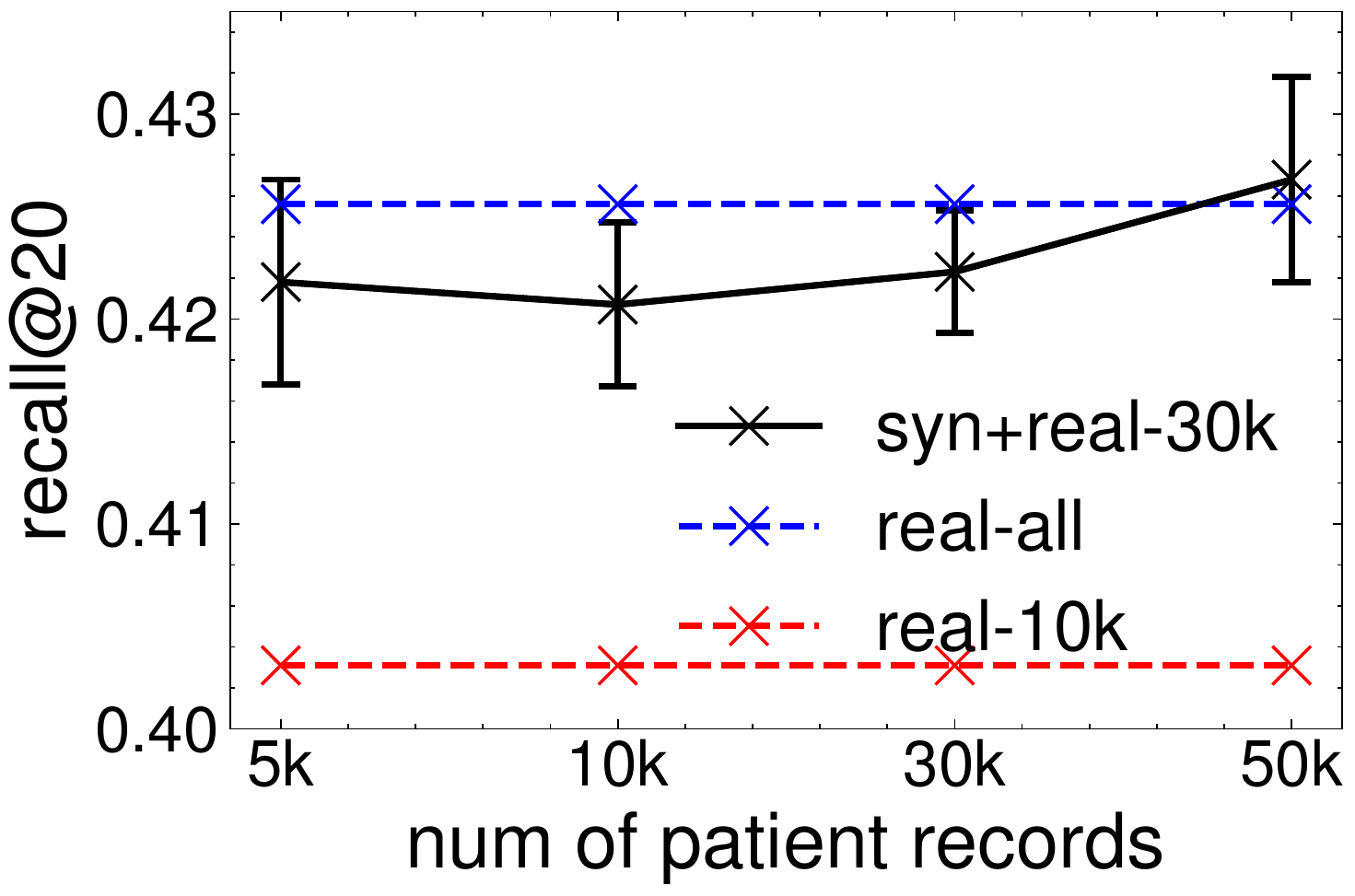}
    \caption{Recall@20. \label{fig:syn_real_30k_rec_20}}
  \end{subfigure}
\caption{Recall of the predictive model on the test set with varying input data size: \emph{syn+real-30k} indicates the model trained on the \emph{hybrid of synthetic \& 30k real data}; \emph{real-30k/all} indicate trained on 30k/all real data. \label{fig:syn_real_30k}}
\vspace{-1em}
\end{figure}

\subsection{Q3. Synthetic EHRs Utility}
We aim to measure the utility of synthetic data when we develop predictive models on top of them. We compare LSTM models on $\mathcal{D}_S$ and $\mathcal{D}_1$ with multilabel prediction for diagnosis events similar to the setting in \cite{choi2016doctor}. In particular, we design two experiments: (1) train LSTM on fully synthetic data and compare its performance with the one trained on real data; (2) train LSTM on a mixture of synthetic data and real data where the synthetic data is regarded as data augmentation.

\noindent \textbf{Fully synthetic data.} We test the LSTM performance on 5k, 10k, 30k, and 50k synthetic patient records. For comparison, the model performance on 5k and 10k real records are also tested. Results are shown in Fig. \ref{fig:syn_rec}. For recall@10 in Fig. \ref{fig:syn_rec_10}, we can observe that though 10k synthetic records are not comparable to 5k real records, 30k synthetic records can reach a better performance than 10k real records. On the other hand, for recall@20 in Fig. \ref{fig:syn_rec_20}, we surprisingly find that 5k synthetic records achieve the same performance as the 5k real records. With more synthetic records involved, the 50k synthetic records-based LSTM outperforms its counterpart on 10k real records at last. This experiment demonstrates that synthetic EHRs by \method are sufficient to support healthcare applications. It is expected to achieve comparable performance by synthetic data as the real data.

\noindent \textbf{Hybrid synthetc-real data.} In Fig. \ref{fig:syn_real_10k}, we randomly sample 10k real data from $\mathcal{D}_1$ and combine them with different sizes of synthetic data from $\mathcal{D}_S$. We find that the model trained on the augmented hybrid data has obvious advantages over its counterpart on the real data. With more synthetic records involved, the model gains better performance. This demonstrates the utility of synthetic data used as augmentation in low-resource cases. Besides, from Fig. \ref{fig:syn_real_10k} we identify this hybrid data is still inferior to the model trained on all real records. So we are curious about how many synthetic and real data we need to \textit{outperform} this seemingly performance upper bound. In other words, can we beat the real data with the synthetic data?

\begin{figure}[t]
  \begin{subfigure}[t]{0.23\textwidth}
    \includegraphics[width=\textwidth]{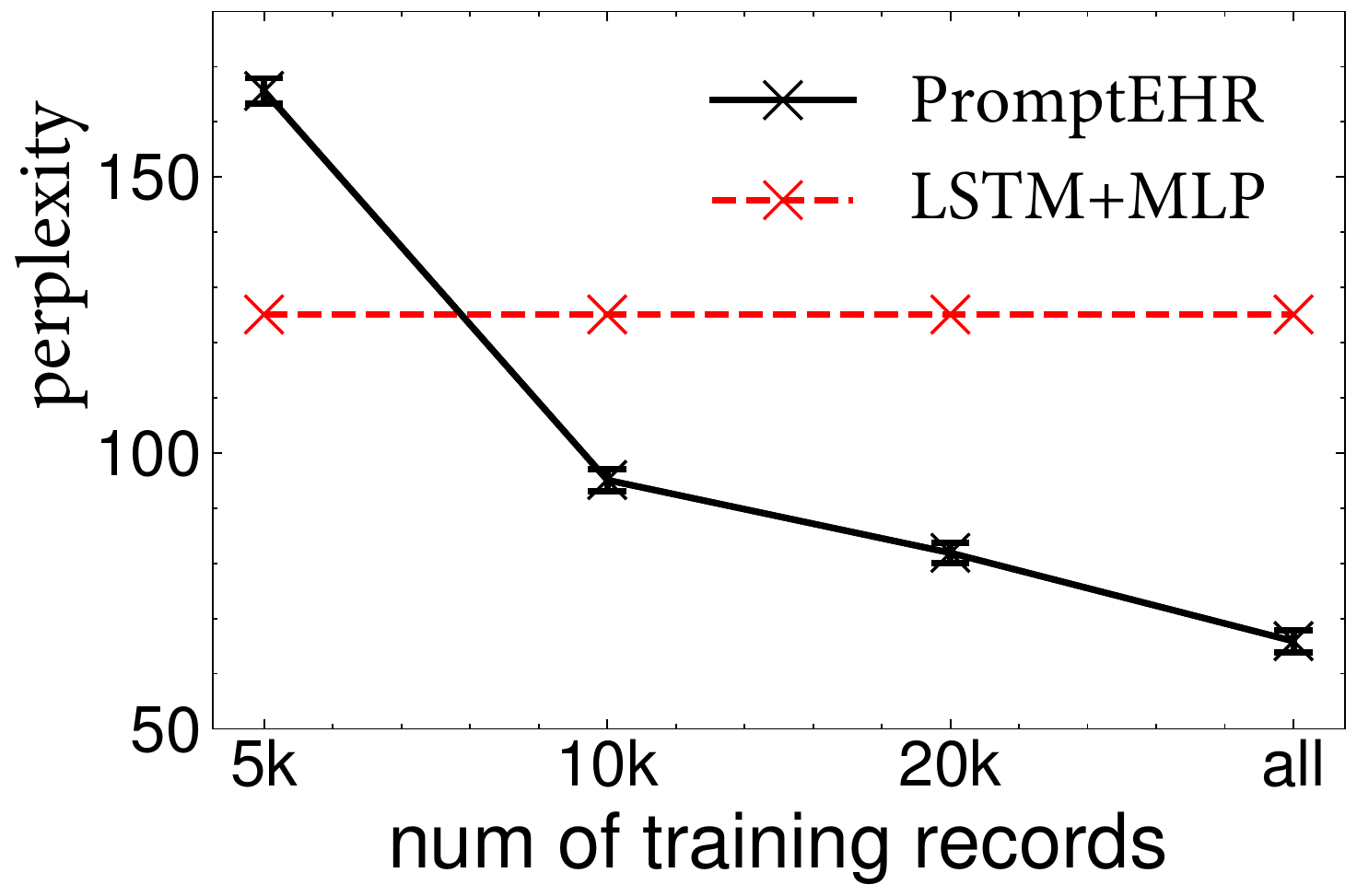}
    \caption{Perplexity (\texttt{lpl}). \label{fig:train_record_tpl}}
  \end{subfigure}
  \hfill
  \begin{subfigure}[t]{0.23\textwidth}
    \includegraphics[width=\textwidth]{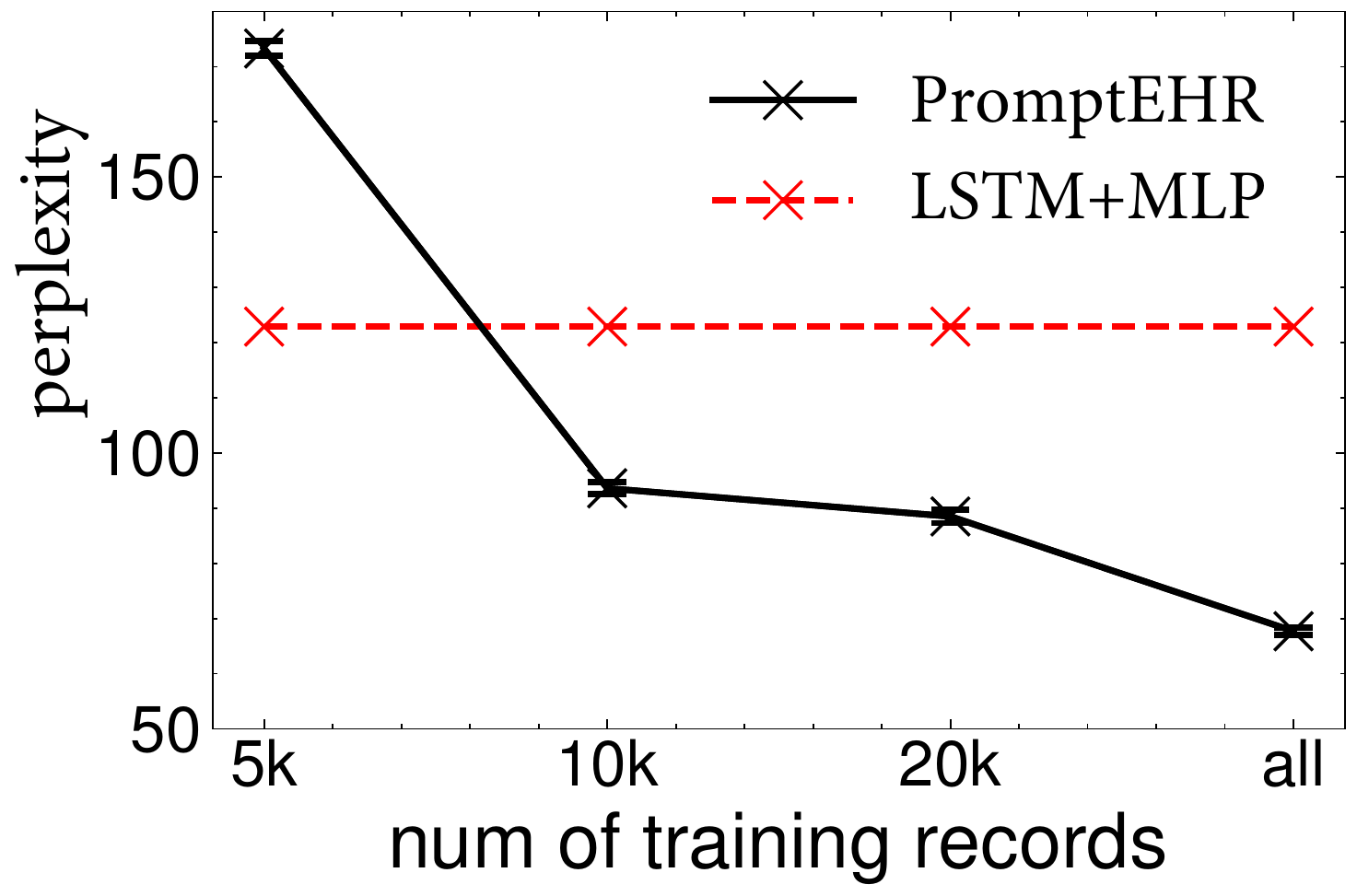}
    \caption{Perplexity (\texttt{mpl}). \label{fig:train_record_spl}}
  \end{subfigure}
\caption{Black solid lines show the spatial and temporal perplexities of \method with regard to varying input training record sizes. Red dotted lines show the \texttt{lpl} and \texttt{mpl} of baseline LSTM+MLP trained on all training records ($\sim$40k). \label{fig:train_record_tpl_spl}}
  \vspace{-1em}
\end{figure}

We conduct the next experiment where 30k real data is combined with synthetic data. Note that we have around 40k real training records in total. Results are shown in Fig. \ref{fig:syn_real_30k}. It can be seen that 50k synthetic records plus 30k real records train better models than on all the real data.

\subsection{Q4. Quality w.r.t. Training Size}
In practice, the original data source to be shared might be in limited size, which elicits a question on how much the generation quality of \method is influenced by the size of the training cohort. To answer this question, we sampled 5k, 10k, and 20k patient records from the training set and testify the perplexity of the learned \method. Results are illustrated by Fig. \ref{fig:train_record_tpl_spl}. We plot the performance of the baseline LSTM+MLP method trained on all real training records ($\sim$40k) in red dotted lines for comparison. It shows that \method trained on 5k training records has worse generation quality than the baseline. When additional 5k records are involved, \method not only outperforms the LSTM baseline but also all other baselines reported in Table \ref{tab:ppl}, which demonstrates that \method is amenable to low resources and superior than the baselines.

\subsection{Case Study}
We demonstrate two use cases of \method: \textit{generating from scratch} (Table \ref{tab:case_from_scratch}) and \textit{generating by completion} (Table \ref{tab:case_from_completion}). While previous works handle the former, only \method handles the completion setting because it makes flexible conditional generation based on either patient features or previous events. In Table \ref{tab:case_from_completion}, our model begins from all diagnosis of one patient and then generates labtests via cross-modal imputation. Then, we randomly sample one procedure and let the model impute all the remaining procedures based on diagnosis and the labtests. Iteratively applying this strategy yields diverse and realistic EHRs via conditional generation. We provide explanations of the two synthetic records in Appendix \S \ref{appx:case_study}.

\section{Conclusion}
In this paper, we study how to leverage real EHRs to train a prompt learning based generative language model for synthetic EHRs generation, namely \method. Unlike previous EHRs generation methods, \method is able to learn from and generate heterogeneous EHRs. To evaluate its performance, we draw the idea of perplexity from the text generation literature and propose two perplexity measures: spatial and temporal perplexity. Experiments on MIMIC-III data demonstrates the quality of generated EHRs are better than the baselines. The synthetic data provides both utility and privacy for downstream healthcare applications.

\section*{Limitations}
This work seeks to generate synthetic records hence avoid sharing sensitive personal electronic healthcare records for the development of machine learning models. In our experiments, we find the generated synthetic records by \method are invulnerable to two adversaries: membership inference and attribute inference. However, there is still possibility that there exists some more advanced attacking methods which can take the advantage of synthetic records. Obviously we cannot exhaust all adversaries for empirical privacy evaluation. In this viewpoint, it is promising to investigate theoretic-guaranteed EHRs generation approach. For instance, we may draw the idea of differential privacy to enhance the current method to provide a complete privacy protection.

\bibliography{custom,anthology}
\bibliographystyle{acl_natbib}

\appendix
\clearpage

\section{Case Study} \label{appx:case_study}
The first case was generated from scratch (Table \ref{tab:case_from_scratch}), it describes a patient who goes into ICU because of a cesarean. During the operation, a test of Hematocrit should be conducted to ensure blood loss of the patient within the safe range. In the second visit, the patient suffers from a bacteria infection. The patient then receives a series of lab tests regarding the inflammation. And spinal tap is performed to help cure serious infections. Antibiotic drugs, e.g., Ampicillin Sodium and Gentamicin, are used to cure the patient. It can be seen that the generated events all center around the same topic (liveborn) and the longitudinal and cross-modal connections are coherent.

The second case was generated based on a real patient EHR by leveraging flexible imputation functions of \method (Table \ref{tab:case_from_completion}). The model scans through the record in time order. For each modality in a visit, we randomly choose to keep all events, remove all events, or remove a part at random. The imputed events are marked red. For example, in visit-1, the model takes the diagnosis codes with prompts as inputs and generates the lab tests. Then, the generated lab tests are involved in the input with prompts. In addition, the procedure 'Enteral infusion of nutrition' is also kept in the inputs. The model then generates the remaining procedures in this visit. This process repeats until reaches visit-6 where the real EHR ends. 

In general, the events in the second case are coherent under the topic of pneumonia and heart failure. The patient is diagnosed as suffering from pneumonia due to bacteria with many complications like a hemorrhage of gastrointestinal tract, heart failure, and pulmonary collapse. At the same time, procedures like the enteral infusion of nutrition, insertion/replacement of endotracheal tube, and temporary tracheostomy are all included to maintain the patient's life regarding his/her nutrition and breath. Besides this visit, the remaining synthetic visits are also reasonable: he/she gets diagnoses regarding heart failure, respiratory diseases, stomach disorders, etc., which all correspond to relevant issues appearing in the first visit. These two cases offer an intuitive demonstration of the effectiveness of \method in generating realistic EHRs, especially when we take the advantage of multiple imputation functions to generate rather realistic EHRs based on real EHRs, which was hardly mentioned in previous works.

\begin{table}[t]
  \centering
  \caption{A synthetic patient generated by \method from scratch. \textit{ICD\_abc} indicates the first three digits represented by ICD code of the event.}
  \resizebox{!}{0.5\linewidth}{
    \begin{tabular}{cp{18em}}
    \hline
    \multicolumn{1}{c}{\multirow{3}[2]{*}{\textbf{Visit-1}}} &  \underline{\textbf{Diagnosis}}: Liveborn \bigstrut[t]\\
          &  \underline{\textbf{Labtest}}: Hematocrit \\
          &  \underline{\textbf{Procedure}}: Prophylactic vaccination \bigstrut[b]\\
    \hline
    \multicolumn{1}{c}{\multirow{4}[2]{*}{\textbf{Visit-2}}} &  \underline{\textbf{Diagnosis}}: Streptococcus infection, Extreme immaturity, Perinatal infection, Neonatal jaundice, Liveborn \bigstrut[t]\\
          &  \underline{\textbf{Labtest}}: Anion Gap, Bands, Base Excess, Bilirubin, Total, Chloride, Eosinophils, Hematocrit, Hemoglobin, Lymphocytes, MCH, MCHC, MCV, Monocytes, Platelet Count, Potassium, Red Blood Cells, Sodium, pCO2, pH, pO2 \\
          &  \underline{\textbf{Drug}}: Ampicillin Sodium, Heparin Sodium (Preservative Free), NEO*IV*Gentamicin, NEO*PO*Ferrous Sulfate Elixir, Send 500mg Vial, Syringe (Neonatal) *D5W* \\
          &  \underline{\textbf{Procedure}}: Biopsy of spinal cord \bigstrut[b]\\
    \hline
    \end{tabular}%
    }
  \label{tab:case_from_scratch}
  \vspace{-1em}
\end{table}%

\begin{table}[t]
  \centering
  \caption{A synthetic patient generated by \method based on a real patient record. The imputed events are marked yellow. For demonstration, we cut the events after the fifth for each visit due to the space limit. }
  \resizebox{!}{\linewidth}{
    \begin{tabular}{cp{24em}}
    \hline
    \multicolumn{1}{c}{\multirow{3}[2]{*}{\textbf{Visit-1}}} & \underline{\textbf{Diagnosis}}: Pneumonia, Hematemesis, Heart failure, Emphysema \bigstrut[t]\\
          &  \underline{\textbf{Labtest}}: \colorbox{yellow}{Leukocytes}, \colorbox{yellow}{Urea Nitrogen}, \colorbox{yellow}{Calcium}, \colorbox{yellow}{Ketone} \\
          &  \underline{\textbf{Procedure}}: Enteral infusion of nutrition, \colorbox{yellow}{Insertion of airway}, \colorbox{yellow}{Replace tracheostomy tube}, \colorbox{yellow}{Temporary tracheostomy} \bigstrut[b]\\
    \hline
    \multicolumn{1}{c}{\multirow{3}[2]{*}{\textbf{Visit-2}}} &  \underline{\textbf{Diagnosis}}: \colorbox{yellow}{Heart failure}, \colorbox{yellow}{Respiratory conditions}, \colorbox{yellow}{Tracheostomy status}, \colorbox{yellow}{Stomach disorder} \bigstrut[t]\\
          &  \underline{\textbf{Labtest}}: Urine Appearance, Yeast, \colorbox{yellow}{Platelet Count}, \colorbox{yellow}{Calculated Total CO2} \\
          &  \underline{\textbf{Procedure}}: Biopsy of bronchus, Replace gastrostomy tube, \colorbox{yellow}{Invasive mechanical ventilation}, \colorbox{yellow}{Infusion of nesiritide} \bigstrut[b]\\
    \hline
    \multicolumn{1}{c}{\multirow{3}[2]{*}{\textbf{Visit-3}}} &  \underline{\textbf{Diagnosis}}: Pneumonia, Mechanical complication, \colorbox{yellow}{Pulmonary manifestations}, \colorbox{yellow}{Disorders of urinary tract} \bigstrut[t]\\
          &  \underline{\textbf{Labtest}}: INR(PT), Epithelial Cells, RBC, \colorbox{yellow}{Urine Appearance} \\
          &  \underline{\textbf{Procedure}}: Insertion of airway, \colorbox{yellow}{Enterostomy}, \colorbox{yellow}{Lysis of peritoneal adhesions}, \colorbox{yellow}{Lung biopsy} \bigstrut[b]\\
    \hline
    \multicolumn{1}{c}{\multirow{3}[2]{*}{\textbf{Visit-4}}} &  \underline{\textbf{Diagnosis}}: Mechanical complication, Hodgkin's paragranuloma, \colorbox{yellow}{Pressure ulcer}, \colorbox{yellow}{Heart failure} \bigstrut[t]\\
          &  \underline{\textbf{Labtest}}: \colorbox{yellow}{Urine Color}, \colorbox{yellow}{Urobilinogen}, \colorbox{yellow}{Bands}, \colorbox{yellow}{Urea Nitrogen} \\
          &  \underline{\textbf{Procedure}}: \colorbox{yellow}{Infusion of nesiritide}, \colorbox{yellow}{Endoscopy of small intestine}, \colorbox{yellow}{Gastrostomy}, \colorbox{yellow}{Replace tracheostomy tube} \bigstrut[b]\\
    \hline
    \multicolumn{1}{c}{\multirow{4}[2]{*}{\textbf{Visit-5}}} &  \underline{\textbf{Diagnosis}}: Urethra disorder, Attention to tracheostomy/gastrostomy, \colorbox{yellow}{Pneumonia}, \colorbox{yellow}{Heart failure} \bigstrut[t]\\
          &  \underline{\textbf{Labtest}}: MCH, \colorbox{yellow}{Bacteria}, \colorbox{yellow}{Lymphocytes}, \colorbox{yellow}{Calculated Total CO2} \\
          &  \underline{\textbf{Drug}}: Fluticasone Propionate 110mcg, \colorbox{yellow}{SW}, \colorbox{yellow}{Bisacodyl}, \colorbox{yellow}{Iso-Osmotic Dextrose} \\
          &  \underline{\textbf{Procedure}}: \colorbox{yellow}{Replace tracheostomy tube}, \colorbox{yellow}{Heart cardiac catheterization}, \colorbox{yellow}{Enteral infusion of nutrition} \bigstrut[b]\\
    \hline
    \multicolumn{1}{c}{\multirow{4}[2]{*}{\textbf{Visit-6}}} &  \underline{\textbf{Diagnosis}}: Pneumonia, Heart failure, \colorbox{yellow}{Endomyocardial fibrosis}, \colorbox{yellow}{Mechanical complication} \bigstrut[t]\\
          &  \underline{\textbf{Labtest}}: pH, Epithelial Cells, \colorbox{yellow}{WBC}, \colorbox{yellow}{Protein} \\
          &  \underline{\textbf{Drug}}: Neutra-Phos, \colorbox{yellow}{Mirtazapine}, \colorbox{yellow}{Fluconazole}, \colorbox{yellow}{SW} \\
          &  \underline{\textbf{Procedure}}: Invasive mechanical ventilation, \colorbox{yellow}{Airway infusion}, \colorbox{yellow}{Monitoring of cardiac output}, \colorbox{yellow}{Lung biospy} \bigstrut[b]\\
    \hline
    \end{tabular}%
    }
  \label{tab:case_from_completion}%
  \vspace{-1em}
\end{table}%
\end{document}